\documentclass[conference,9pt]{IEEEtran}

\IEEEoverridecommandlockouts

\usepackage{cite}
\usepackage{amsmath,amssymb,amsfonts}
\usepackage{algorithmic}
\usepackage{graphicx}
\usepackage{fancyhdr}
\usepackage{textcomp}
\usepackage{epstopdf}
\usepackage[dvipsnames]{xcolor}
\usepackage{paralist}
\usepackage[inline]{enumitem}
\usepackage{caption}
\usepackage{multirow}
\usepackage{booktabs}
\usepackage{makecell}
\usepackage{tabu, booktabs}
\usepackage{float}
\usepackage{subcaption}
\usepackage{pifont}
\usepackage[a4paper, total={184mm,239mm}]{geometry}
\def\BibTeX{{\rm B\kern-.05em{\sc i\kern-.025em b}\kern-.08em
    T\kern-.1667em\lower.7ex\hbox{E}\kern-.125emX}}

\ExplSyntaxOn
\DeclareExpandableDocumentCommand{\convertlen}{ O{cm} m }
 {
  \dim_to_decimal_in_unit:nn { #2 } { 1 #1 } cm
 }
\ExplSyntaxOff
\usepackage{hyperref}
\usepackage{gnuplottex}
\usepackage{tikz}

\usetikzlibrary{matrix,calc}

\newcommand*\circled[1]{\tikz[baseline=(char.base)]{
            \node[shape=circle,fill,inner sep=0.5pt] (char) {\textcolor{white}{#1}};}}

\newcommand\myshade{85}

\definecolor{royalblue}{RGB}{65, 105, 225}
\definecolor{darkgreen}{RGB}{0, 100, 0}
\definecolor{darkred}{RGB}{139, 0, 0}
\definecolor{darkviolet}{RGB}{148, 0, 211}

\colorlet{linkColor}{violet}
\colorlet{citeColor}{YellowOrange}
\colorlet{urlColor}{Emerald}

\hypersetup{
  linkcolor  = darkviolet!\myshade!black,
  citecolor  = citeColor!\myshade!black,
  urlcolor   = urlColor!\myshade!black,
  filecolor  = darkgreen!\myshade!black,
  colorlinks = true,
}

\begin{document}

\title{OmniBoost: Boosting Throughput of Heterogeneous Embedded Devices under Multi-DNN Workload}

\author{
    Andreas Karatzas and Iraklis Anagnostopoulos \\
    School of Electrical, Computer and Biomedical Engineering, Southern Illinois University, Carbondale, U.S.A.\\
    \{andreas.karatzas,iraklis.anagno\}@siu.edu
}

\maketitle

\thispagestyle{fancy}
\fancyhead{} 
\rhead{} 
\lhead{} 
\chead{{\small Accepted for publication at the 60th Design Automation Conference (DAC 2023). \textcopyright2023 IEEE.}} 
\renewcommand{\headrulewidth}{0.4pt} 

\begin{abstract}

Modern Deep Neural Networks (DNNs) exhibit profound efficiency and accuracy properties. 
This has introduced application workloads that comprise of multiple DNN applications, raising new challenges regarding workload distribution. 
Equipped with a diverse set of accelerators, newer embedded system present architectural heterogeneity, which current run-time controllers are unable to fully utilize. 
To enable high throughput in multi-DNN workloads, such a controller is ought to explore hundreds of thousands of possible solutions to exploit the underlying heterogeneity. 
In this paper, we propose OmniBoost, a lightweight and extensible multi-DNN manager for heterogeneous embedded devices. 
We leverage stochastic space exploration and we combine it with a highly accurate performance estimator to observe a $\times 4.6$ average throughput boost compared to other state-of-the-art methods. 
The evaluation was performed on the HiKey970 development board.

\end{abstract}

\begin{IEEEkeywords}
Deep Neural Networks, Concurrent Deep Learning Scheduling, Heterogeneous Architectures, Edge Inference, DNN Performance Prediction, Embedded Deep Learning
\end{IEEEkeywords}

\setlist{nosep}

\section{Introduction}

With the recent advancements in the area of machine learning, Deep Neural Networks (DNNs) have become the driving force for modern embedded devices. In particular, a wide variety of embedded applications such as digital assistants, object detection, and virtual/augmented reality services heavily rely on DNNs~\cite{kwon2021heterogeneous}. As such devices are generally characterized by limited computing capabilities and thus, platform heterogeneity has prevailed as a solution to increase system performance. In particular, heterogeneity can be classified into performance and functional. Performance heterogeneity refers to the integration of processing elements with different performance but the same Instruction Set Architecture (iso-ISA cores). Functional heterogeneity refers to the integration of specialized processing elements with different ISA, such as embedded GPUs.

However, this architectural heterogeneity of modern embedded devices has introduced new challenges regarding the execution of DNNs~\cite{kang2020scheduling,hsieh2019surf}, as current deep learning frameworks do not utilize all the underlying heterogeneity efficiently. They mostly utilize either the CPU or the GPU, but not both. This is due to the complexity of the programming frameworks and the challenges of data communication between the different computing components. Additionally, the common belief that GPUs are better platforms for DNN execution does not always stand true in the embedded domain. Actually, \emph{Meta (Facebook) showed that Android devices utilize CPUs for the excution of DNNs and only a small fraction of inferences are mapped on the GPUs due to programmability and performance limitations~\cite{wu2019machine}}. Hence, in order to exploit the underlying heterogeneity, deep learning frameworks must be evolved and utilize the full potential of all computing resources.

Moreover, many modern applications simultaneously employ multiple and different DNNs to support sophisticated and complex services~\cite{kwon2021heterogeneous,cox2021masa}. To that end, heterogeneous embedded devices must execute multi-DNN workloads consisting of various DNNs, each one comprising different amounts of operations and layers. This behavior imposes additional scheduling challenges to the underlying hardware, as the resources must be collaboratively utilized to support the required quality of service. Intuitively, mapping multiple DNNs only on computationally strong processing elements (e.g., GPUs) saturates these units~\cite{hsieh2019case}, while utilizing low performing components (e.g., LITTLE CPUs) significantly affects system throughput~\cite{wang2019high}. 

To balance the overall multi-DNN workload of the system, schedulers must
\begin{inparaenum}[(i)]
    \item exploit the inter-layer parallelism properties of DNN models~\cite{nichols2021survey} to design efficient pipelines at run-time and
    \item partition the layers of the DNNs over the underlying computing components in an optimal way.
\end{inparaenum}
Considering though that modern DNNs consist of multiple layers and the number of heterogeneous components increases, collaboratively finding a multi-DNN workload with the optimal partition for each DNN and the corresponding computing device is very complex due to the vast design space~\cite{kang2020scheduling}. Therefore, the goal is to design a scheduler that can exploit computational patterns and yield optimal solutions with high probability. Additionally, the scheduler must be lightweight, in order to reduce the decision latency at run-time, and extensible, in order to support multiple DNNs without manual tuning.

In this work, we present OmniBoost, a framework that utilizes both performance and functional heterogeneity of embedded devices to increase system throughput via DNN partitioning. In particular, OmniBoost uses distributed embedding vectors to correlate the performance of each DNN layer with the computing characteristics of the underlying components. The core idea of OmniBoost is the Monte Carlo Tree Search (MCTS) algorithm that can efficiently explore a large search space. It also utilizes a lightweight estimator to evaluate the given space and output the near optimal solutions with high probability. We deployed and tested OmniBoost on the HiKey970 development board. Experimental results show that in heavy multi-DNN workloads, while other methods yield solutions that saturate the system,
OmniBoost finds mappings that evenly distribute the given workload leading to a $\times 4.6$ average throughput speedup.~\\
\noindent
\textbf{Overall, the main contributions of this work are as follows:}
\begin{itemize}
    \item We propose a lightweight and extensible multi-DNN scheduler that utilizes DNN layer partitioning to boost throughput on heterogeneous embedded devices.
    \item We utilize Monte Carlo Tree Search (MCTS) for managing design space exploration under budget constraints.
    \item We design a run-time workload throughput estimator based on distributed embedding vectors and residual connections for managing decisions at inference.
    \item We ported and evaluated the proposed framework on HiKey970 board for various multi-DNN workload scenarios. 
\end{itemize}

\section{Motivation}\label{sec:motivation}

This section contains a motivational example that shows why synergistically utilizing both CPU and GPU can boost throughput on embedded devices under multi-DNN workload. We also show how big the decision space is and the necessity for an efficient exploration mechanism.

For our motivational example, we utilized the HiKey970 board which features a Mali-G72 MP12 GPU and big.LITTLE CPUs with a quad-core A73 running at 2.36GHz and a quad-core Cortex-A53 at 1.8GHz, and we selected four widely used DNNs:
\begin{inparaenum}[(i)]
    \item AlexNet~\cite{alexnet}, 
    \item MobileNet~\cite{mobilenet},
    \item VGG-19~\cite{vgg}, and 
    \item SqueezeNet~\cite{squeezenet}.
\end{inparaenum}
Following the common scheduling approach, we mapped all of them on the GPU and we recorded the average inferences per second for all DNNs. 
Then, we created 200 different set-ups, in which we randomly split the layers of the DNNs between the big CPU and the GPU. An example set-up is:
\begin{inparaenum}[(i)]
    \item AlexNet: first 4 layers on GPU, the remaining on big CPU;
    \item MobileNet: first 10 layers on big CPU, the remaining on GPU;
    \item VGG-19: first 15 layers on GPU, the remaining on big CPU; and
    \item SqueezeNet: first 18 layers on GPU, the last one on LITLLE CPU.
\end{inparaenum}
For each one of the randomly generated set-ups, we also recorded the average inferences per second.

\begin{figure}[htbp]
    \centering
    \includegraphics[width=0.46\textwidth]{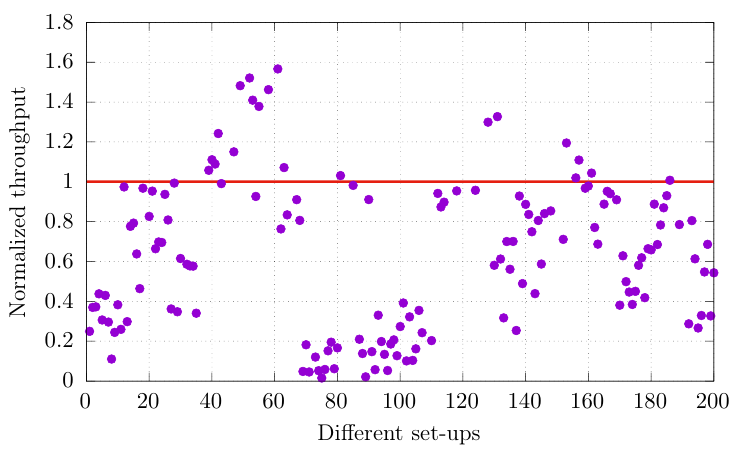}
    \caption{Normalized throughput of four concurrent DNNs under different set-ups. For each set-up we randomly split the layers between CPU and GPU. Baseline is the case in which all DNNs are executed on the GPU.}
    \label{fig:motivation}
\end{figure}

Figure~\ref{fig:motivation} shows the normalized performance of each set-up. The baseline is considered to be the case in which all the layers of the DNNs are executed on the GPU. We observe that even though the baseline achieves higher throughput than most of the set-ups, there are many cases in which splitting the layers across the system's computing components increases the performance. In our example, the best set-up increased the performance by up to $60\%$.

Regarding the discovery of the best set-up and the exploration cost, the number of possible combinations makes a greedy search infeasible. For this particular example only, the total number of combinations is $C_3(84) = {84 \choose 3} \approx 95,000$, where $84$ is the total number of DNN layers scheduled and 3 the number of different computing components. Nevertheless, this number of combinations corresponds only to these particular DNN models. This means that if we expand our dataset and consider more DNN architectures, the sum of those combinations yield a design space in the order of millions. Eventually, scheduling multiple DNNs applications across on heterogeneous systems renders a well-know NP-hard problem as shown in~\cite{kang2020scheduling}.
\section{Related Work}\label{sec:related}

The authors in~\cite{wang2019high} propose Pipe-it, a framework that utilizes micro-benchmarks to create a layer performance prediction model and build DNN pipelines. However, Pipe-it is limited on CPU-only deployments. In this work, we consider pipeline execution on GPUs as well. In~\cite{hsieh2019case}, the authors show that the DNN throughput can be increased by distributing the computational workload across the system's computing components. Similarly, CNNDroid~\cite{salar2015cnndroid} accelerates the execution of a CNN application by mapping the convolutional layers, which typically pose a heavy computational load, to the GPU leaving the rest, less computationally intensive layers, to be executed on the CPU. However, in both methods the process followed is static and the GPU workload can quickly reach saturation point while managing multiple CNN applications.
In our work, we model the partition point for each DNN  as a variable yielding a solution space to be searched for the optimal DNN mapping.
SURF~\cite{hsieh2019surf} proposes dynamic layer splitting as a way to optimize throughput by heuristically compiling a DNN pipeline.
However, it suffers from scalability issues since the defined solution space increases exponentially for every new DNN added to the dataset.
We resolve this problem by stochastically exploring the solution space and most importantly parameterizing the computational budget for the search process.
Another framework, namely RSTensorFlow~\cite{alzantot2017rstensorflow}, utilizes RenderScript~\cite{renderscript} to exploit both inter- and intra-layer parallelism for modern DNNs.
Nonetheless, RSTensorFlow does not consider the workload of each computing component while searching for an optimal DNN mapping. Our method, OmniBoost, addresses this challenge by synergistically considering the whole set of the DNNs in the workload to yield a solution.
The authors in~\cite{kwon2021heterogeneous} present a framework that greedily assigns layers to system's computing components. 
However, the employed software architecture uses a trial-and-error method, which introduces scalability and performance concerns due to its space exploration inefficiency.
AI-MT~\cite{baek2020multi}, another framework that manages multi-DNN workloads, utilizes a latency estimation model to balance the workload across the system's computing components. However, the performance estimator in AI-MT statically evaluates the expected workload with respect to the memory requirements of each model layer. This is inefficient in the case of heterogeneous embedded systems due to the limited computational resources which constitute static performance estimators obsolete. 
The authors in~\cite{spantidi2022targeting} also proposed a solution for multi-DNN workloads on heterogeneous embedded systems comprised of NPUs. 
However, inter-layer parallelism properties of DNNs are not utilized, thus exhibiting an imbalance in workload distribution.
PRISM~\cite{das2022real} proposes a scheduling framework for heterogeneous Neuromporphic Systems-on-Chip (NSoCs). However, the heuristic backbone used to explore the design space requires multiple model graph evaluations. This is impractical in our case, since instead of transaction orders, we would have to perform forward passes, which would result in a huge run-time overhead.
Moreover, the authors in~\cite{han2019mosaic} propose a linear regression model to automate layer splitting and DNN mapping onto heterogeneous embedded systems. Although intuitively efficient, this approach is built upon the assumption that the execution time of DNN layers is linearly correlated to the dimensions of input matrices.
However, this assumption is not held in the case of multi-DNN scheduling, thus yielding sub-optimal results.
An improvement in terms of performance estimation is proposed in~\cite{kang2020scheduling}. 
The authors developed a scheduler using a genetic algorithm that addresses multi-DNN workloads.
However, this approach is computationally expensive and lacks scalability.
Also, it yields solutions that involve frequent data transfers due to the large amount of pipeline elements in those solutions which introduce unnecessary delay due to data dependencies. 
We resolve the first challenge by introducing a throughput estimator which can quickly model a performance policy due to its few trainable parameters. 
Secondly, we reinforce our framework to choose mappings that introduce minimal pipeline stages and hence remove the undesired delay.
Similarly, the authors in~\cite{cox2021masa} introduce MASA, a scheduling algorithm for multi-DNN workload across heterogeneous embedded systems. 
\begin{figure*}[htbp]
    \centering
    \resizebox{0.82\textwidth}{!}{\includegraphics[clip,width=\textwidth]{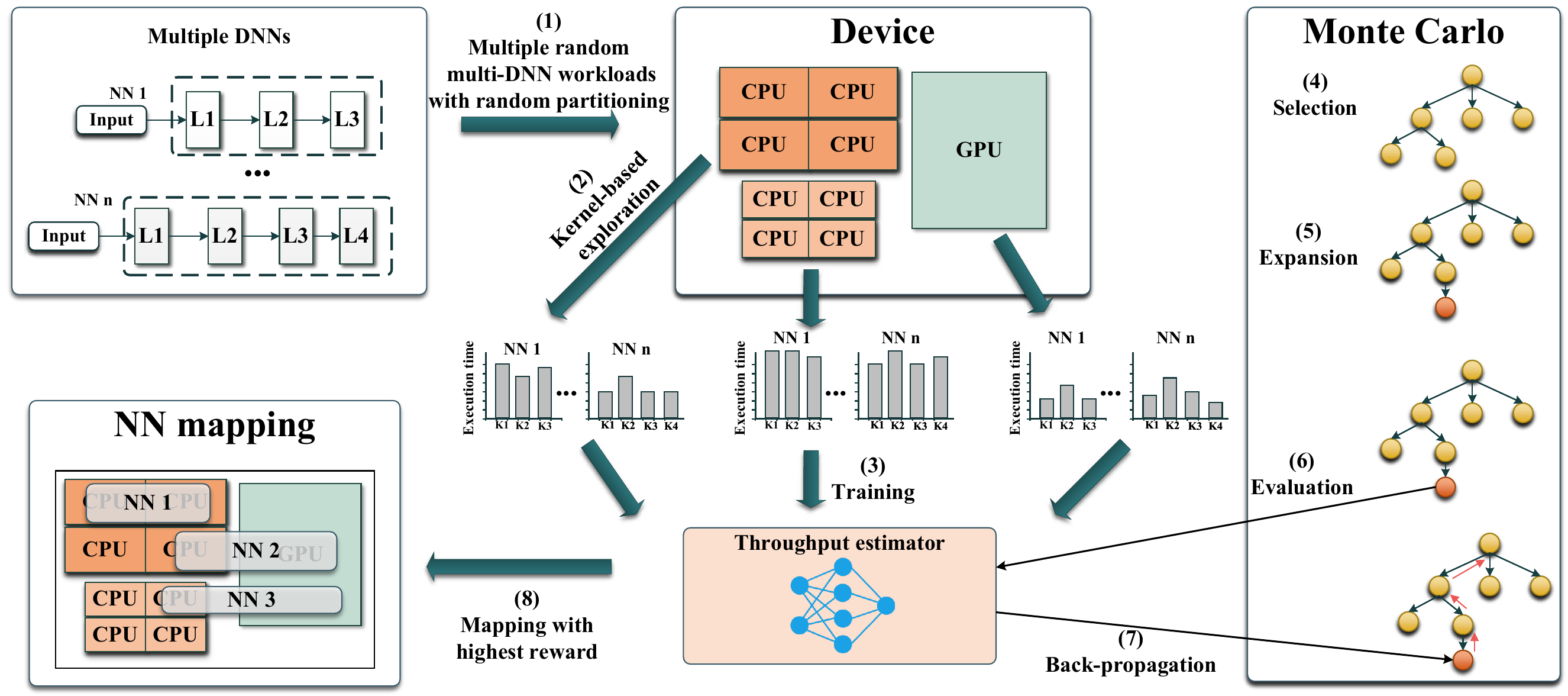}}
    \caption{A high-level overview of the proposed framework.}
    \label{fig:OmniBoost-high-level-overview}
\end{figure*}
However, the framework architecture disregards the minimization of pipeline stages, thus hindering solutions with undesired data dependencies and delay.

Our key differences from the state-of-the-art are manyfold:
\begin{inparaenum}[(i)]
    \item we enable concurrent and efficient multi-DNN execution using a highly scalable workload representation that leverages the properties of distributed embedding vectors;
    \item we introduce a highly accurate workload performance estimator that can generalize with only several dataset samples;
    \item OmniBoost is designed to be robust to new DNN models added on top of the existing dataset;
    \item using a probabilistic approach, we efficiently prune the vast search space.
\end{inparaenum}

\section{Proposed Scheduling Framework}\label{sec:framework}

OmniBoost is a framework that utilizes inter-layer parallelism in DNN applications to build efficient pipelines that execute concurrently and boost the overall system throughput on heterogeneous embedded systems.
Figure~\ref{fig:OmniBoost-high-level-overview} depicts the overview of the proposed approach.
The proposed software architecture is compiled of:
\begin{inparaenum}[(i)]
    \item the estimator model, which is a CNN that is trained to evaluate any given mapping, and
    \item the Monte Carlo Tree Search (MCTS) algorithm.
\end{inparaenum}
In essence, OmniBoost is comprised of an exploration and a ranking mechanism, i.e. the MCTS algorithm and the estimator respectively.

\subsection{The Distributed Embeddings Tensor}\label{subsec:dist-embeddings}

In this section, we formulate the input to our throughput estimator.
To that end, we compile a distributed embeddings tensor representing the performance of each DNN layer when executed on each computing component of the device (i.e., GPU, big CPU, and LITTLE CPU). 
To capture that information, we perform a kernel-based exploration by recording the execution time of each kernel on the device.
Therefore, the performance of each DNN layer with respect to a computing component is defined as:
\begin{equation}
    B_{\alpha}^{l} = \sum_{k \in l}^{k} b_{\alpha}^{k}
\end{equation}
where $B$ is the execution time of layer $l$ running on the computing component $\alpha$ (i.e., GPU, big CPU, and LITTLE CPU), and $b$ is the execution time of the kernel $k$ in layer $l$.
By adopting kernel-based granularity, we focus on the individual building blocks of each layer, which offers greater adaptability when incorporating new DNN models. This granularity allows for the identification and optimization of specific computational patterns across diverse models, enabling the framework to accommodate various DNN architectures with minimal adjustments. Consequently, this approach streamlines the integration process and reduces the effort required to adapt the framework to new DNN models, resulting in a more versatile and efficient system.
Each DNN model is comprised of several layers, therefore we compile a performance vector for each DNN in our dataset as follows:
\begin{equation}
    p_{\alpha}^{m} = \begin{pmatrix} B_{\alpha}^{1} & B_{\alpha}^{2} & \dots & B_{\alpha}^{n} \end{pmatrix}
\end{equation}
where $n$ is the number of layers in DNN $m$, and $p$ the performance vector corresponding to a that specific DNN $m$ while running on the computing component $\alpha$. 
The result is a performance matrix $P$ compiled by the performance vectors $p$ of all DNNs for each computing component $a$:
\begin{equation}
    P_{\alpha} = \begin{pmatrix}
        \begin{bmatrix}
            \vdots \\
            p_{\alpha}^{1} \\
            \vdots
        \end{bmatrix} &
        \begin{bmatrix}
            \vdots \\
            p_{\alpha}^{2} \\
            \vdots
        \end{bmatrix} &
        \dots &
        \begin{bmatrix}
            \vdots \\
            p_{\alpha}^{M} \\
            \vdots
        \end{bmatrix}
    \end{pmatrix}
\end{equation}
where $M$ is the number of DNN models in our dataset.
However, every model has a different number of layers. 
To overcome this structural difficulty, we employ zero-padding so that the number of elements for each vector $p_{\alpha}^{m}$ is equalized.
Since in our case we utilize the Hikey970 board with $3$ computing components (i.e., GPU, big CPU, and little CPU), the resulting structure is a compilation of three performance matrices stacked together to essentially build one tensor with $3$ slices, which is referenced as $U$.
Finally, the term distributed~\cite{embeddings} refers to the nature of information distribution, which is not concentrated around a single point, but scattered across the entire tensor.

The resulting tensor holds the information for all DNNs in our dataset and every computing component of our platform.
Henceforth, we need to find a mechanism that extracts specific information with respect to a defined system workload.
This can be achieved with mask tensors, i.e., boolean tensors that are convolved with tensor $U$, to yield a new tensor that has non-zero elements only where a computing element has assigned workload.

\subsection{The Performance Estimator}\label{subsec:estimator-cnn}

Following the compilation of a masked embedding tensor as described in~\ref{subsec:dist-embeddings}, we design a processing module that handles the throughout estimation of the corresponding system workload. 
We develop a lightweight ResNet9-based CNN performance estimator~\cite{resnet} with only $20,044$ trainable parameters, enabling faster training and effective pattern discovery in small datasets. Our model incorporates GELU~\cite{gelu}, a Gaussian Error Linear Unit activation function, instead of the original ReLU~\cite{relu}, which has shown overall improvements in both model convergence and accuracy.

The performance estimator module's purpose is to predict the average throughput for each computing component under any given workload, accounting for both data transfer delays within the pipeline and computational latency in each unit.
For our case, this means that there will be $3$ neurons in the output layer.
We do not utilize any activation function in the output layer, since the model is trained to solve a regression problem. 
In other words, the purpose of the proposed CNN is not to classify the input data. Contrary, the goal is to transform its input into a vector that represents the expected normalized system throughput of the given heterogeneous embedded system with respect to the assigned workload as defined by that input. 

\begin{figure}[htbp]
    \centering
    \resizebox{0.5\textwidth}{!}{\includegraphics[clip,width=\textwidth]{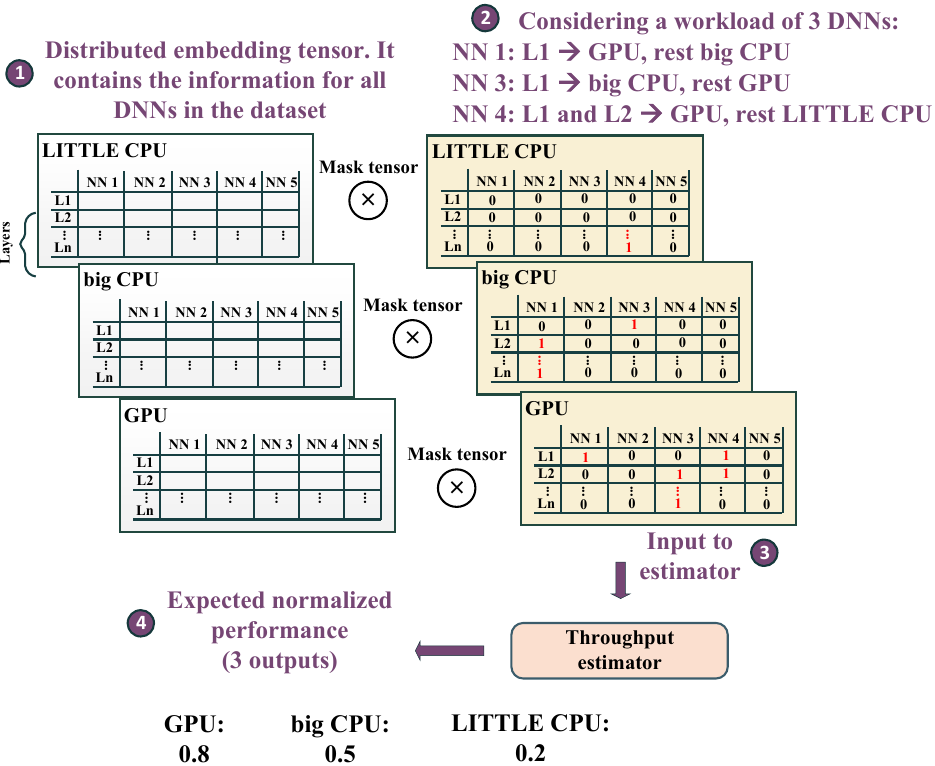}}
    \caption{Throughput estimator processing flow example.}
    \label{fig:example}
\end{figure}

Figure~\ref{fig:example} shows an example of the processing flow of our throughput estimator. In step \circled{1}, we fetch the distributed embedding tensor, which holds the benchmark data per DNN layer for all the computing components of the system. In particular, each cell contains the normalized execution time of each layer per computing component.
In step \circled{2}, we mask the embedding tensor with respect to the queried workload. We perform an element-wise multiplication between the mask tensor and the embedding tensor, filtering out elements which correspond to DNN layers that are not scheduled on that particular computing component. In step \circled{3}, we propagate the masked tensor to the throughput estimator. Finally, in step \circled{4}, we observe the output of the performance estimator. The output is a vector, which holds the normalized predicted throughput for each computing component with respect to the queried mappings.

\subsection{The MCTS Algorithm}\label{subsec:mcts}

Even though we trained the performance estimator module, it is designed as a regression mechanism. 
In other words, it is unable to render a mapping given a set of DNNs by itself. 
One approach is to exhaustively feed all the candidate solutions spanned by that DNN set combined with the computing components of the given heterogeneous embedded system.
However, the number of valid mappings that need to be evaluated are tens of millions.
Therefore, it is imperative to find an efficient space exploration framework.
To that end, the MCTS algorithm~\cite{mcts-recent}, a heuristic that tackles the curse of dimensionality by approaching the given problem in a tree-based decision structure, under a given computational budget~\cite{alphago}.

To deploy MCTS, we define an environment composed of \textbf{states}, \textbf{actions}, and \textbf{rewards}. 
MCTS progressively compiles a solution by stochastically expanding a \textbf{state} tree. 
Each state is associated with an expected \textbf{reward}, and the MCTS algorithm aims to maximize that metric and thus return the most efficient decision.
As in every game environment, winning and losing states must be well defined. 
We define a winning state as the state where the MCTS has correlated every DNN layer to a computing component for all the models in a given set.
To reinforce MCTS to reach a winning state, we associate the winning one with an exceptionally high reward.
Similarly, losing states are mostly characterized by inefficiencies in the resulting pipeline structure.
Specifically, if MCTS yields a solution with a number of pipeline stages higher than $x$, which is the number of computing components on a given heterogeneous embedded system, the solution is labeled as inefficient marking the corresponding tree node losing.
The reason why $x$ is set to be equal to the number of computing components is to avoid redundant pipeline stages, thus minimizing data transfers and undesired performance delays.
In every step, the MCTS algorithm selects a tree node based on its expected reward and expands it.
A node expansion means the random exploration of a path under the selected node.
The randomly selected path ends in a leaf node.
Afterwards, we use our trained performance estimator to evaluate the selected action trajectory and estimate a reward.
The output is then propagated to the compiled tree structure and the whole process is repeated until a computational constraint is met.
Finally, the MCTS algorithm fetches (i) the candidate state with the highest expected reward and (ii) a decision is made based on that elite state's associated action trajectory.
In our case, each node can have at most $3$ child nodes, since there are $3$ computing components on our heterogeneous system.
leaf nodes receive static rewards based on winning or losing states, while non-leaf nodes are evaluated by the performance estimator, which returns the expected system throughput as a reward in the defined environment.

The last important topic is to define the environment \textbf{actions}.
We define $3$ available actions and each of those actions correspond to a specific computing component.
At first the MCTS chooses a computing component for the first DNN layer. 
However, for implementational purposes, we assume the whole DNN to be handled by that component.
Afterwards, the MCTS is queried to map each DNN layer to one of the $3$ computing components, starting from the second one and gradually reaching the last one.
After scheduling a DNN model, the MCTS algorithm fetches the next DNN and repeats the process until all DNNs are scheduled.
The order of DNNs, in any defined set, is not important because those DNN models are eventually mapped to run concurrently.

\section{Experimental results}\label{sec:evaluation}

In this section, we assess the efficiency of our proposed framework by performing an in-depth evaluation over various multi-DNN workloads on HiKey970 development board.
HiKey970 features a Mali-G72 MP12 GPU and big.LITTLE CPUs with a quad-core A73 running at 2.36GHz and a quad-core Cortex-A53 at 1.8GHz.
Although the HiKey970 features an NPU, it was not employed in this study due to compatibility issues with the utilized compute library.
OmniBoost is implemented in PyTorch~\cite{pytorch} and OpenAI Gym~\cite{openai}, while OpenCL and ARM Compute Library~\cite{armcl} were used for executing the DNNs on the board and splitting the partitions.

\begin{figure}[htbp]
    \centering
    \includegraphics[width=0.46\textwidth]{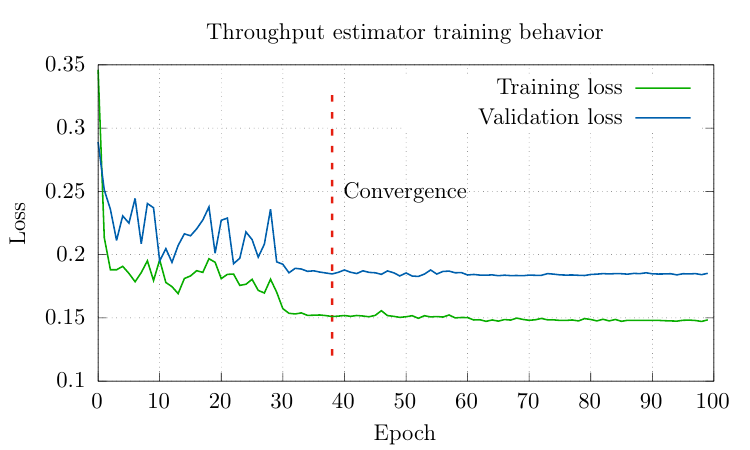}
    \caption{Training and validation loss curves for our novel CNN estimator.}
    \label{fig:training-stats}
\end{figure}

Regarding the training of the throughput estimator during the design-time, we created 500 workloads, consisting of random mixes ranging from 1 up to 5 concurrent DNNs. The DNNs that we used to create the random mixes are:
\begin{inparaenum}[(i)]
    \item AlexNet~\cite{alexnet}
    \item MobileNet~\cite{mobilenet}
    \item ResNet-34~\cite{resnet}
    \item ResNet-50
    \item ResNet-101
    \item VGG-13~\cite{vgg}
    \item VGG-16
    \item VGG-19
    \item SqueezeNet~\cite{squeezenet}
    \item Inception-v3~\cite{inceptionv3}, and
    \item Inception-v4~\cite{inceptionv4}.
\end{inparaenum}
Each mix was randomly distributed across the computing components of the device, in order to create samples with different pressure on the computing components. Due to the vast design space arising from three computing components and multiple valid partition points per DNN, these workloads possess high probability of uniqueness. Next, each of these mixes was rendered as a single tensor to feed the performance estimator module of OmniBoost. The target output of our performance estimator is the average system throughput measured with respect to those generated mixes. 
To circumvent numerical pitfalls in our dataset, we incorporate two preprocessing methods into our architecture. The first standardizes the dataset output to address large variations and non-uniform distribution, while the second normalizes the output vector elements to values between $0$ and $1$.
These two preprocessing layers mitigate potential issues arising from the raw data, such as poor convergence, slow learning, or suboptimal model performance. Standardization alleviates issues related to large variations and non-uniform distribution, ensuring consistent data input for our performance estimator. Meanwhile, normalization addresses the problem of varying magnitudes by scaling the dataset.
The dataset was split in 400 training and 100 validation samples. Since the problem solved by the proposed neural network is of a regression nature, we used L1-loss as a criterion.
We also trained our model using L2-loss function, but it proved to be too aggressive in some cases, thus resulting in sub-optimal model weights.
The CNN estimator was trained for 100 epochs on a NVIDIA 1660 Ti GPU, which took under a minute.
The training stats of our model can be examined in Figure~\ref{fig:training-stats}.

\subsection{Throughput comparison}\label{subsec:throughput}

For performance evaluation, we constructed multiple random mixes of concurrent DNNs. In particular, we created mixes consisting of 3, 4, and 5 concurrent DNNs.
We also tried mixes with 6 concurrent DNNs, but the pressure on the memory controller and the computing resources exceeded the board's computational capabilities thus rendering it unresponsive.

For each mix, we measured the average throughput defined as follows: $T = \frac{\sum_{1}^{M} \frac{INF_i}{sec}}{M}$, where $\frac{INF}{sec}$ is the processed inferences per second of each DNN in the mix, and $M$ is the total number of DNNs in it.
We compared the average throughput of OmniBoost against three other approaches:
\begin{inparaenum}[(i)]
    \item the common scheduling approach, which maps all the given workload to the GPU, the highest performing device (also used as the baseline in this work);
    \item a linear regression-based algorithm followed in MOSAIC~\cite{han2019mosaic}; and 
    \item the Genetic Algorithm (GA) approach presented in~\cite{kang2020scheduling}.
\end{inparaenum}

\begin{figure}[htbp]
    \centering
    \subfloat[Normalized average throughput for five different mixes. Each mix consists of 3 concurrent executing DNNs, selected randomly.\label{fig:mix3}]{%
      \includegraphics[width=0.46\textwidth]{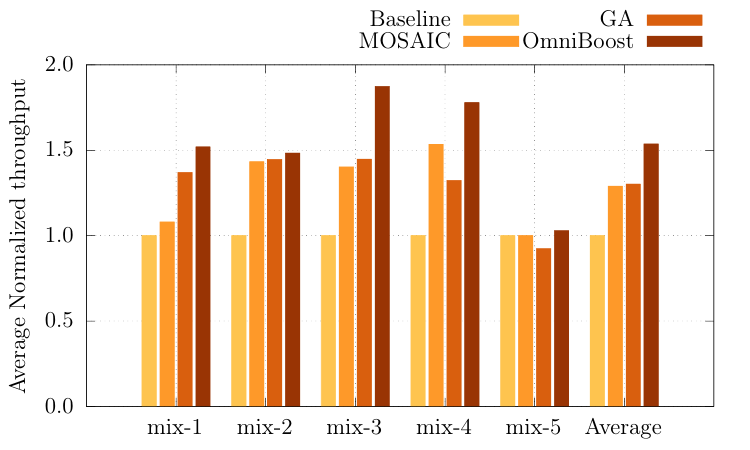}%
    }
    
    \subfloat[Normalized average throughput for five different mixes. Each mix consists of 4 concurrent executing DNNs, selected randomly.\label{fig:mix4}]{%
      \includegraphics[width=0.46\textwidth]{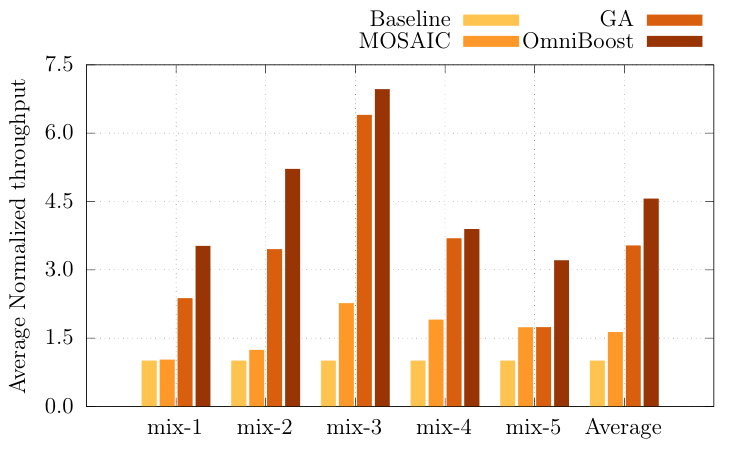}%
    }
    
    \subfloat[Normalized average throughput for five different mixes. Each mix consists of 5 concurrent executing DNNs, selected randomly.\label{fig:mix5}]{%
      \includegraphics[width=0.46\textwidth]{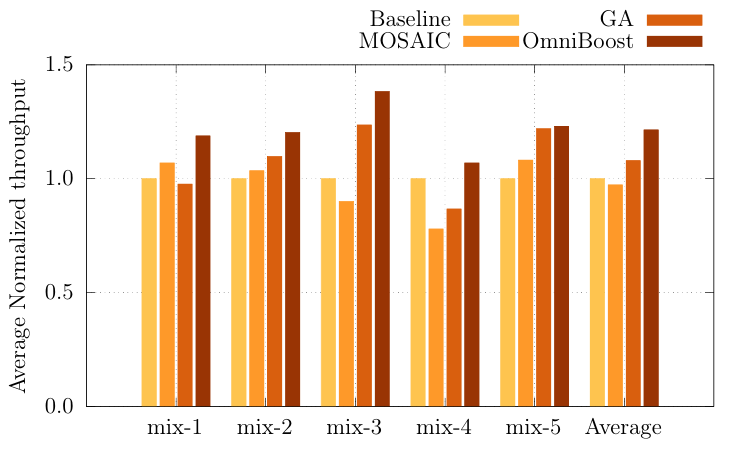}%
    }
    \caption{Average system throughput for (a) the baseline scheduling approach, (b) the MOSAIC framework, (c) the GA-based framework and (d) OmniBoost.}
    \label{fig:experiments}
\end{figure}

We set the computational budget of MCTS algorithm to 500 and the search depth to 100, which resulted in fast solution space exploration and thus an overall performance optimization.
It's worth mentioning that the operators utilized in the Genetic Algorithm actually damage the candidate solutions rather than contributing towards the overall population converge.
For example, mutation might introduce extra latency to an elite chromosome due to the newly introduced pipeline stages.
That's why we have integrated an optimization layer that heuristically merges redundant pipeline stages.
Furthermore, the GA needs retraining for every new queried workload.
To the best of our knowledge, OmniBoost is the first framework that addresses the multi-DNN scheduling problem without retraining.

Figure~\ref{fig:mix3} depicts the normalized average throughput for five random mixes consisting of 3 DNNs each. 
On average, OmniBoost achieves on average $54\%$, $19\%$, and $18\%$ higher throughput compared to the baseline, the MOSAIC, and the GA method accordingly. 
Running 3 DNNs concurrently, does not push the computing resources of the board to its limit. 
This is shown in mix-5, where all frameworks yield similar solutions in terms of throughput. 
This is because the computing components of the device are able to handle the workload and they have not reached yet a saturation point. 
In particular, mix-5 consists of lightweight DNNs such as AlexNet, VGG-13, and MobileNet.

In Figure~\ref{fig:mix4}, we show the normalized average throughput for five random mixes of 4 DNNs each. 
Again, OmniBoost achieved on average $\times 4.6$ and $\times 2.83$ compared to the baseline approach and the MOSAIC framework accordingly.

The gains are considerably higher than before due to the fact that the baseline approach and the MOSAIC framework overloaded the device's GPU. Contrary, the GA and OmniBoost better distribute the overall DNN workload and hence yield better solutions, with OmniBoost presenting a substantial improvement compared to the GA by $23\%$.

Finally, Figure~\ref{fig:mix5} depicts the normalized average throughput for five random mixes of 5 DNNs each. 
Overall, MOSAIC falls behind the baseline approach by $2.7\%$, while the GA and OmniBoost surpass the baseline approach by $7\%$ and $22\%$ respectively. 
Despite the properties of the GA and OmniBoost to efficiently exploit the system's underlying heterogeneity, the queried workload is not as efficiently managed as before. 
This is due to fact that running 5 DNNs simultaneously on the board, overloads all the computing resources, saturating throughput.  
As aforementioned, we also tried mixes with 6 concurrent DNNs, but the overall workload too heavy for the limited computing resources of the board, thus making it unresponsive.

\subsection{Run-time performance evaluation}

Regarding the execution time, the baseline method is the most efficient among all due to its simplicity as it places all DNNs on the GPU without any decision overhead. 
However, this approach fails to utilize the system's underlying heterogeneity, thus yielding the worst throughput results, as we showed in Figure~\ref{fig:experiments}.

Regarding MOSAIC, it executes a single query to the trained linear regression model. To that end, the actual inference time is really low ($\sim 1 sec$).
However, the overhead of MOSAIC lies in the very time consuming factor of the data collection process.
MOSAIC is trained using more than $14,000$ data points, which constitutes a notable time interval. Additionally, MOSAIC produces sub-optimal results. Thus, the fast decision yields sub-optimal results and trade-offs significant throughput gains, as shown in Figures~\ref{fig:mix3}-\ref{fig:mix5}.

Regarding the GA method, it requires retraining for each queried workload, which highly increases its run-time response. 
Another time-consuming characteristic of the GA solution is the optimization layer that aids in lowering the pipeline stages.
While this layer optimizes the GA's overall performance, it injects a large overhead, thus hindering the overall framework response time. GA needed approximately 5 minutes for each mix presented in Section~\ref{subsec:throughput}.

Finally, OmniBoost has a constant number of iterations which is experimentally set to 500. This implies 500 queries to the performance evaluation module, which predominantly impacts the decision latency. Even though this is seemingly a large number of queries, the overall framework performance is satisfactory, due to the low number of trainable parameters that compiles our performance evaluation module. In our experiments, OmniBoost found an efficient mapping in approximately 30 seconds.
It's worth noting that the budgetary constraints can be adjusted for any use-case scenario, thus introducing extra flexibility. Overall, OmniBoost showcases the best trade-off between run-time performance and throughput optimization thereby assuming the most efficient framework of all.

\section{Conclusion}

Modern heterogeneous embedded systems that execute Deep Neural Networks (DNNs) impose grave challenges on the interface of workload management. 
Those challenges are augmented in cases where the given workload is comprised of multiple DNN applications.
To leverage all the available computational resources, the system's installed run-time controller must utilize the underlying heterogeneity and distribute the defined workload.
To that end, this paper presents OmniBoost, a lightweight and extensible multi-DNN manager for heterogeneous embedded devices that can efficiently distribute any given multi-DNN workload.
Our extensive experimental evaluation, shows that our framework can optimally utilize the device's computing components to boost system throughput significantly.
It is noteworthy that our framework achieves $\times 4.6$ average throughput boost while keeping the run-time performance overhead down to a minimal.
\section{Acknowledgment}

This research has been funded in part by the Consortium for Embedded Systems at SIUC.

\footnotesize
\bibliographystyle{IEEEtran}
\bibliography{ref}

\end{document}